\documentclass[letterpaper, 10 pt, conference]{ieeeconf}  

\IEEEoverridecommandlockouts                              

\usepackage{graphicx}
\usepackage{amsmath}
\usepackage{amssymb}
\usepackage{booktabs}
\usepackage{bm} 
\usepackage{bbm}
\usepackage{dsfont} 
\usepackage{comment}
\usepackage{url}

\title{\LARGE \bf
ProMi: An Efficient Prototype-Mixture Baseline for Few-Shot Segmentation with Bounding-Box Annotations
}

\author{Florent Chiaroni$^{1}$, Ali Ayub$^{2}$, Ola Ahmad$^{1}$
\thanks{$^{1}$
Thales, CortAIx-Lab, Montreal, QC, Canada. {\tt\small florent.chiaroni@thalesdigitalsolutions.ca, ola.ahmad@thalesgroup.com}}%
\thanks{$^{2}$
Concordia University, Montreal, QC, Canada {\tt\small ali.ayub@concordia.ca}}%
}

\begin{document}

\maketitle
\thispagestyle{empty}
\pagestyle{empty}

\begin{abstract}
In robotics applications, few-shot segmentation is crucial because it allows robots to perform complex tasks with minimal training data, facilitating their adaptation to diverse, real-world environments. However, pixel-level annotations of even small amount of images is highly time-consuming and costly. In this paper, we present a novel few-shot binary segmentation method based on bounding-box annotations instead of pixel-level labels. We introduce, ProMi, an efficient prototype-mixture-based method that treats the background class as a mixture of distributions. Our approach is simple, training-free, and effective, accommodating coarse annotations with ease. Compared to existing baselines, ProMi achieves the best results across different datasets with significant gains, demonstrating its effectiveness. Furthermore, we present qualitative experiments tailored to real-world mobile robot tasks, demonstrating the applicability of our approach in such scenarios. Our code: \url{https://github.com/ThalesGroup/promi}.
\end{abstract}

\section{INTRODUCTION}
\label{sec:intro}

\noindent
Recent advances in robotics, particularly through the integration of computer vision, have greatly enhanced robots' ability to perceive and interact with their environment. Key innovations such as object detection \cite{7780460, Ren2015FasterRT, Cheng2024YOLOWorld}, scene understanding \cite{Premebida2018IntelligentRP}, and image segmentation \cite{Kirillov2023SegmentA}, as well as vision-based real-time path planning and navigation \cite{7219438, Uppal_2024_CVPR}, now enable robots to interpret and analyze their surroundings more effectively, leading to more efficient interactions and sophisticated decision-making. In this work, we focus on image segmentation as it is a fundamental task for robot scene understanding, safe navigation including obstacle avoidance, and numerous other robotic applications.

Image segmentation involves assigning a label to each pixel in an image that corresponds to objects of interest. Recent deep learning models have excelled in this task, often surpassing human performance \cite{7410480}. Their success is largely attributed to advancements in architecture design \cite{krizhevsky2012imagenet}, \cite{simonyan2014very}, \cite{he2016deep}, \cite{vaswani2017attention}, \cite{DosovitskiyB0WZ21}, enabling them to effectively learn from large, annotated datasets. In real-world systems, such as autonomous robots, obtaining high-quality annotations from large datasets is often impractical due to the significant costs associated with human annotation and data collection. These processes are both time-consuming and expensive, making it challenging to gather the labeled data necessary for training these models. Few-shot learning strategies are particularly valuable in such scenarios, as they significantly reduce the cost and effort involved in manual labeling \cite{9523110}. However, in image segmentation, even labeling a small number of images at the pixel level remains highly time-consuming and costly. This has motivated the development of methods that avoid the reliance on pixel level annotations, such as weakly supervised few-shot learning methods \cite{han2023learning}.

The problem becomes more crucial when deploying segmentation models in autonomous systems that must interact with their environment in real time. The model may encounter new object classes unseen during training, as well as a high degree of uncertainty due to environmental changes. These systems must adapt and segment novel objects on the fly. However, most existing solutions overlook operational constraints and focus solely on the segmentation task.

In this paper, we present an efficient solution to address the current challenges. Specifically, we propose a novel weakly supervised few-shot binary segmentation approach that: 1) requires only a few bounding-box annotations, significantly reducing the annotation burden, and 2) allows the segmentation model to adapt quickly to new objects or environments. Our approach, named ProMi (Prototype-Mixture), uses an efficient prototype-based technique and models the background class as a mixture of distributions, enabling better handling of complex background variations and improving performance in difficult settings. We present experiments on common datasets used in few-shot segmentation tasks, demonstrating that ProMi is significantly more robust to noisy weak annotations and complex backgrounds when compared to all baselines, including prototype-based few-shot methods. We also present qualitative experiments on datasets tailored to real-world mobile robot tasks.

In short, we make the following key contributions:
\begin{itemize}
    \item Problem formulation: transitioning from bounding-box annotations to partially noisy pixel-wise annotations. 
    \item ProMi: a novel prototype mixture-based approach specifically designed for few-shot binary segmentation with bounding-box annotations.
    \item Extensive comparative experiments, including adaptations of prototype-based few-shot classification methods to the segmentation tasks and evaluations with a foundation model feature extractor.
\end{itemize}

\section{RELATED WORK}

\subsection{Few-Shot Learning}
Few-Shot Learning (FSL) addresses the challenge of learning new tasks with a few training samples, by transferring the general knowledge learned through large datasets to novel tasks. Meta-learning is one of the common techniques in FSL, where an ML model is trained on how to learn from only a few samples~\cite{snell2017prototypical,finn2017model,sung2018learning}. ProtoNet~\cite{snell2017prototypical} is one of the most commonly used meta-learning techniques, which leverages the prototypes (mean feature vectors) to perform nearest neighbor classification. Later works in FSL~\cite{chen2019closerfewshot,tian2020rethinking,mandi2022effectiveness} showed that meta-learning does not generalize well to new tasks that might be different from base training data. Simply using frozen pre-trained feature extractors and linear or prototypical classifiers lead to much better generalizability, robustness, and overall performance~\cite{chen2019closerfewshot}. Inspired by this, we propose a simple prototype-based few-shot segmentation method that is more robust and computationally efficient than prior works in the field. Our method alternates between data assignment and parameter updates like k-means \cite{macqueen1967some}, \cite{lloyd1982least} and Expectation-Maximization (EM) \cite{dempster1977maximum} but can also dynamically increase the number of prototypes.

\subsection{Few-Shot Segmentation}
Few-shot segmentation (FSS) techniques consider the challenge of segmenting new images while learning from only a few training examples. Similar to few-shot classification, a significant number of works in FSS are prototypical learning methods. Inspired by ProtoNet~\cite{snell2017prototypical}, several FSS methods consolidate the support set information into a single~\cite{zhang2020sg,wang2019panet,cao2024prototype,zhang2022mask,liu2022intermediate,boudiaf2021few} or multiple prototypes~\cite{lang2022beyond,yang2020prototype,liu2022learning,zhang2022feature,okazawa2022interclass,wang2022adaptive,zhang2021self,liu2022dynamic} followed by feature comparison or aggregation. In most prototype-based methods, segmentation is performed by pixel-wise classification, where each pixel is classified as background or foreground based on the distance between the prototypes and the feature vector associated with the pixel, whereas some develop adaptive classifiers with prototypes serving as the initial weights of the classifier~\cite{boudiaf2021few}. While these techniques are effective, they require pixel-level annotations, which remain highly costly and time-consuming, particularly in real-time robotic applications.

\subsection{Weakly-Supervised Few-Shot Segmentation}
Instead of relying on strongly labelled pixel-level annotations, weakly-supervised few-shot segmentation (WSFSS) leverages weaker labels, including bounding boxes or scribbles \cite{wang2019panet, zhang2019canet, han2023learning}, image-level class labels \cite{raza2019weakly, kang2023distilling}, word embeddings combined with image-level class labels \cite{siam2020weakly}. WSFSS methods that depend on image-level class labels have proven less effective at segmenting novel objects~\cite{raza2019weakly}. In addition, they often provide inaccurate segmentation when dealing with complex backgrounds. Moreover, many WSFSS methods use meta-learning techniques and require a large number of segmentation masks during meta-training \cite{han2023learning}, making them less efficient. In contrast, our approach neither relies on meta-learning nor requires segmentation masks.

\section{PROBLEM FORMULATION}
\label{sec_problem_formulation}

\subsection{Target challenge: Few-Shot Binary Segmentation with Bounding-box Annotations}

In the few-shot binary segmentation task we address in this article, the objective is to segment the foreground object in a query image using a support set annotated with bounding-boxes rather than pixel-level masks. Despite this coarse annotation, pixel-level predictions are required for the query set. Formally, the task is defined as follows:

\noindent \textbf{Support Set:} Let the support set \(\mathcal{S} = \{(I_i^s, \mathcal{B}_i^s)\}_{i=1}^N\) consist of \(N\) support images \(I_i^s \in \mathbb{R}^{H \times W \times C}\) and their corresponding sets of bounding-box annotations \(\mathcal{B}_i^s\). Here, \(H\) and \(W\) are the height and width of the images, and \(C\) is the number of color channels. Each bounding-box corresponds to an object belonging to the foreground class of interest. 

\noindent \textbf{Query Set:} The query set \(\mathcal{Q} = \{(I^q, M^q)\}\) consists of a query image \(I^q \in \mathbb{R}^{H \times W \times C}\) and its unknown ground truth binary mask \(M^q \in \{0, 1\}^{H \times W}\). The goal is to accurately predict \(M^q\) by segmenting the query image at the pixel level, determining for each pixel \(I_{xy}^q\) whether it belongs to the foreground (\(M_{xy}^q=1\)) or the background (\(M_{xy}^q=0\)), where \( (x, y) \) denotes the pixel coordinates.

\subsection{Motivation for Our Prototype-Based Strategy}

\textbf{Prototype-based} strategies have proven effective in few-shot image classification \cite{chen2019closerfewshot}, \cite{wang2019simpleshot} due to their simplicity and ability to generalize from limited data. These methods typically involve extracting feature maps from images using a pre-trained encoder, normalizing the features, and then employing a prototype-based linear classifier on top of the features. Prototypes are calculated as the mean of normalized feature vectors for each class, using the labeled images from the support set. This allows for efficient classification by assigning new samples (i.e. query samples) to the nearest prototype.

\textbf{Image segmentation is a classification problem at the pixel-level.} It is worth noting that one could extend such prototype-based strategies to image segmentation by formulating the segmentation task as a pixel-level classification task. Instead of using a 1D feature map for the entire image, segmentation utilizes 2D feature maps with dimensions $\frac{W}{W_p} \times \frac{H}{H_p} \times D$, where each 1D feature vector of size $D$ in this latent space is a representation of a patch of pixels of size $W_p \times H_p$ in the input image space. This could enable a prototype-based model to efficiently perform softmax predictions at the patch-level. Then, these patch-level soft predictions could be resized to the original image dimensions, and applying argmax to each pixel-wise softmax output would yield a binary segmentation mask that aligns with the input image resolution~\cite{wang2019simpleshot,boudiaf2021few}.

\textbf{Limitations of prototype-based segmentation strategies.}
However, we have identified two major limitations for the above mentioned strategy in our context: 
\begin{itemize}
    \item \textbf{Background is a mixture of multiple classes.} A key challenge in segmenting natural images is that using a single prototype~\cite{boudiaf2021few} may fail to capture the diversity of the background class, which often consists of a mixture of different counter-example class distributions. 
    \item \textbf{Coarse bounding-box annotations instead of pixel-level annotations}. Furthermore, in the context of a support set with bounding-box annotations, the labels for the foreground class are inherently noisy. Pixels within bounding boxes include both foreground and background pixels, making the annotation imprecise. This poses a significant challenge for WSFSS techniques that depend on prototypes and bounding boxes, and remains unresolved.
\end{itemize}

To address these limitations and enhance segmentation performance, we propose a prototype-based solution that effectively manages the complexities of diverse background distributions and the noise in foreground labels from bounding-box annotations, as detailed in the next section.

\section{PROPOSED APPROACH}
\label{sec_method}

\noindent
\textbf{Overview.} We present a novel approach, termed \textit{ProMi} (Prototype-Mixture), for few-shot segmentation with bounding-box annotations. It consists of three main steps: (i) First, we encode the support set images to obtain feature maps (Sec. \ref{subsec_feature_extraction}). Concurrently, we convert bounding-box annotations into noisy patch-level labels to exploit them during training (Sec. \ref{subsec_annot_conversion}). (ii) Next, we train our ProMi model using the previously obtained labeled feature maps from the support set. ProMi is tailored to address complex backgrounds and manage noisy support labels (Sec. \ref{subsec_ProMi}), i.e. coarse bounding-box annotations. (iii) Finally, during inference, we encode the query image and use our classifier to generate predictions in the latent space, which are then converted into a segmentation mask matching the input image size (Sec. \ref{subsec_inference}).

\subsection{Feature Extraction} 
\label{subsec_feature_extraction}
At both the training and inference stages, we begin by encoding the input images using a pre-trained encoder to extract spatial feature maps for each image. 
We use a pre-trained encoder \( g_{\bm{\theta}}: \mathcal{I} \rightarrow \mathcal{F} \) to map input images \( I \in \mathcal{I} \subset \mathbb{R}^{H \times W \times C} \) into spatial feature maps \( F \in \mathcal{F} \subset \mathbb{R}^{\frac{H}{H_p} \times \frac{W}{W_p} \times D} \), where \(\bm{\theta}\) represents the encoder parameters, \(H_p\) and \(W_p\) are the embedding reduction ratios for height and width, respectively, and \(D\) is the depth of the feature vectors. Each spatial feature map \( F \) can be represented as a set of feature vectors \( F = \{\bm{f}_m \in \mathbb{R}^{D}\}_{m=1}^{N_p} \), where \( N_p = \frac{H}{H_p} \times \frac{W}{W_p} \). Each feature vector \(\bm{f}_m\) can be considered as the embedding of a corresponding patch of size \(W_p \times H_p\) in the input image.

\subsection{Annotation Conversion: From Bounding-Box to Latent Space-Ready Patch-Level Labels.} 
\label{subsec_annot_conversion}

Given bounding-box annotations in the support set, we directly generate patch-level labels that are suitable for training our prototype-based classifier in the latent space: 

Let \(\tilde{y}_{im}^s \in \{0,1\}\) denote the binary label of the \(m\)-th patch of the image \(I_i^s\) from the support set \(\mathcal{S}\). The patch label \(\tilde{y}_{im}^s\) is set to 1 (noisy positive) if the majority of pixels in the patch are inside any bounding box in \(\mathcal{B}_i^s\), and 0 (true negative) otherwise. As each patch label \(\tilde{y}_{im}^s\) is directly associated with the corresponding feature vector \(\bm{f}_{im}^s\) in the latent space, this conversion enables the creation of the following labeled support set in the latent space:
\[
F^s = \left\{ \bm{f}_{im}^s, \tilde{y}_{im}^s \right\}_{i=1, m=1}^{N, N_p}.
\]

The next step is to train our classifier on $F^s$.

\subsection{ProMi: Prototype mixture-based learning in 
latent space}
\label{subsec_ProMi}

Our proposed classifier, ProMi, operates in the latent space using a prototype-based strategy and cosine similarity for classifying the feature vectors. Given a class of interest, we initially compute one prototype for  the foreground (class of interest) and one for the background. However, these initial prototypes may be suboptimal due to noisy bounding box annotations affecting the foreground and the complex, mixture nature of the background. To address this, ProMi iteratively adds new background prototypes, up to a predefined maximum $K_{max}$, while simultaneously refining the existing prototypes. 
Formally, ProMi consists of a set of prototypes 
\[
W = \{\bm{w}_k\}_{k=0}^{K} = \{\bm{w}_\text{FG}\} \cup \{\bm{w}_\text{BG}^{(k)}\}_{k=1}^{K} \subset \mathbb{R}^D,
\] 
where \(\bm{w}_{0} = \bm{w}_\text{FG}\) is the foreground class prototype, and 
\(\bm{w}_{k} = \bm{w}^{(k)}_\text{BG}\) for \(k \in \{1, \ldots, K\}\) are the background class prototypes. The number of background prototypes, \(K\), starts at 1 and increases iteratively during the refinement process until it reaches \(K_{max}\).

\noindent \textbf{Prototype Initialization.} 
The classifier starts with an initial set of prototypes \(W = \{\bm{w}_k\}_{k=0}^{K=1} = \{\bm{w}_\text{FG}, \bm{w}_\text{BG}^{(1)}\} \subset \mathbb{R}^D\), with \(\bm{w}_{1} = \bm{w}_\text{BG}^{(1)}\) the first background prototype. These prototypes are computed as means of the feature vectors associated with the respective classes:
\[
\bm{w}_\text{FG} = \frac{\sum_{i, m} \tilde{y}_{im}^s 
\bm{f}_{im}^s}{\sum_{i, m} \tilde{y}_{im}^s}, \quad 
\bm{w}_\text{BG}^{(1)} = \frac{\sum_{i, m} (1 - \tilde{y}_{im}^s) \bm{f}_{im}^s}{\sum_{i, m} (1 - \tilde{y}_{im}^s)},
\]
where the summation is over all patches \(m\) and images \(i\) in the support set, i.e. over all feature vectors $\bm{f}_{im}^s$ in $F^s$, with \(\tilde{y}_{im}^s\) indicating whether a feature vector is annotated foreground (1) or background (0). 

\noindent \textbf{Iterative Prototype Refinement.}
To improve classification accuracy, the classifier iteratively refines the prototypes through the following steps:

\begin{enumerate}
    \item \textbf{Hard-label predictions in the latent space:} At each iteration, the classifier calculates the cosine similarity between each feature vector \(\bm{f}_{im}^s\) and all current prototypes in $W$.
    Here, we apply L2 normalization to all feature vectors to ensure they have unit length for the cosine similarity measure.  
    The cosine similarity between a feature vector \(\bm{f}_{im}^s\) and a prototype \(\bm{w}_k\) is then defined as:
    \[
    \cos(\bm{f}_{im}^s, \bm{w}_k) = \bm{f}_{im}^s \cdot \bm{w}_k, \quad \forall k \in \{0, \ldots, K\}.
    \] 
    Hard predictions are then obtained by taking the argmax of these similarities:
    \[
    \hat{y}_{im} = \arg\max_{k} \cos(\bm{f}_{im}^s, \bm{w}_k), \quad \forall k \in \{0, \ldots, K\}.
    \]
    Here, $\hat{y}_{im}=0$ denotes a foreground prediction, while $\hat{y}_{im} \geq 1$ denotes a background prediction \footnote{This maintains consistency with prototype indexes, but differs from class labels (\(\tilde{y}_{im}^s\)), where 0 refers to background and 1 refers to foreground.}.
    
    \item \textbf{Update of Background Prototypes:} Each existing background prototype \(\bm{w}_\text{BG}^{(k)}\) is refined as the mean of the feature vectors assigned to it, as follows:
    \begin{equation*}
    \bm{w}_\text{BG}^{(k)} = \frac{\sum_{i, m} (1 - \tilde{y}_{im}^s
    ) \bm{f}_{im}^s \mathds{1}(\hat{y}_{im} = k)}{\sum_{i, m} (1 - \tilde{y}_{im}^s
    ) \mathds{1}(\hat{y}_{im} = k)},
    \end{equation*}
    for all \(k \in \{1, \ldots, K\}\). Here, \(\mathds{1}(\hat{y}_{im} = k)\) selects the feature vectors \(\bm{f}_{im}^s\) assigned to \(\bm{w}_\text{BG}^{(k)}\). This ensures that distinct prototypes capture different background characteristics.

    \item \textbf{Adding New Background Prototypes:} If false positives are identified, i.e., feature vectors predicted as foreground (\(\hat{y}_{im} = 0\)) 
    but actually belonging to the background (\(\tilde{y}_{im}^s = 0\)), then a new background prototype \(\bm{w}_\text{BG}^{(K+1)}\) is estimated as follows:
    \[
    \bm{w}_\text{BG}^{(K+1)} = \frac{\sum_{i, m} (1 - \tilde{y}_{im}^s) \bm{f}_{im}^s \mathds{1}(\hat{y}_{im} = 0)}{\sum_{i, m} (1 - \tilde{y}_{im}^s) \mathds{1}(\hat{y}_{im} = 0)},
    \]
    corresponding to the mean of all feature vectors identified as false positives during this iteration. This new prototype is added to the set $W$, and the current value of $K$ is incremented by $1$ accordingly.
    \item \textbf{Refinement of the Foreground Prototype:} The foreground prototype \(\bm{w}_\text{FG}\) is refined using feature vectors that are predicted as positive (\(\hat{y}_{im} = 0\)) and correspond to patches labeled as foreground (\(\tilde{y}_{im}^s = 1\)):
    \[
    \bm{w}_\text{FG} = \frac{\sum_{i, m} \tilde{y}_{im
    }^s \bm{f}_{im}^s \mathds{1}(\hat{y}_{im} = 0)}{\sum_{i, m} \tilde{y}_{im
    }^s \mathds{1}(\hat{y}_{im} = 0)}.
    \]
    This refinement step helps the foreground prototype to better capture the distribution of the class of interest despite the noise in the foreground annotations of the support set.
\end{enumerate}

\noindent \textbf{Stopping Criterion.} The proposed prototype-based iterative process continues until either the maximum number of prototypes (\(K_{max}\), a hyperparameter) is reached or no additional false positives are detected. 

\subsection{Inference on the Query set}
\label{subsec_inference}

\textbf{Encoding.} During inference, each query image \( I^q \in \mathcal{Q} \) is first encoded using the pre-trained encoder \( g_{\bm{\theta}} \), producing a spatial feature map \( F^q = g_{\bm{\theta}}(I^q) \), as described in sec. \ref{subsec_feature_extraction}. The feature map is represented as a set of L2-normalized feature vectors \( F^q = \{\bm{f}_m^q \in \mathbb{R}^D \} \), with \( m \in \{1, \ldots, N_{p}\} \).

\textbf{Soft predictions, organized and resized.} The feature vectors \( F^q \) are the inputs to ProMi, our prototype-based classifier, which uses the set of prototypes \( W = \{\bm{w}_k\}_{k=0}^{K} = \{\bm{w}_\text{FG}, \bm{w}_\text{BG}^{(1)}, \ldots, \bm{w}_\text{BG}^{(K)}\} \) previously estimated on the support set (Sec. \ref{subsec_ProMi}). The classifier computes the cosine similarities between each feature vector \(\bm{f}_m^q\) and the prototypes in \( W \), producing the following logits:
\[
\bm{s}_m^k = \cos(\bm{f}_m^q, \bm{w}_k), \quad \forall k \in \{0, \ldots, K\}.
\]
These logits are spatially reorganized into a prediction map of dimensions \(\frac{W}{W_p} \times \frac{H}{H_p} \times (K+1)\). The prediction map is then resized using linear interpolation to match the original width and height of the query image, resulting in a resized prediction map of size \( W \times H \times (K+1) \).

\textbf{Final pixel-wise segmentation.} We then apply \(\arg\max\) on the resized prediction map such that we determine the prototype index with the highest similarity for each pixel. A pixel is classified as background if the highest similarity index corresponds to any of the background prototypes \(\bm{w}_\text{BG}^{(k)}\), and as foreground if it matches the foreground prototype \(\bm{w}_\text{FG}\). In this way, we generate, for each query image, a pixel-level binary segmentation mask $M^q \in \{0, 1\}^{H \times W}$, where \( M_{xy}^q = 1 \) indicates a foreground pixel, and \( M_{xy}^q = 0 \) indicates a background pixel.

\section{EXPERIMENTS}
\label{sec_experiments}

\subsection{Experiments Overview}

\noindent We comprehensively evaluate our approach through distinct sets of experiments to demonstrate its effectiveness across different scenarios. We first report results on standard few-shot segmentation benchmarks using bounding-box annotations for a fair comparison with prior methods (Sec.~\ref{subsec_standard_exps}), followed by experiments with the powerful foundation model Dinov2~\cite{oquab2024dinov} as a feature extractor (Sec.~\ref{subsec_foundation_model_exps}). We then demonstrate the effectiveness of our method on datasets encountered by real-world robots across aerial, ground, and underwater environments (Sec.~\ref{subsec_qualitative_exps}). Finally, we perform an ablation study to analyze the effect of different components of our method (Sec.~\ref{subsec_ablation_study}).

\textbf{Evaluation Metric.} For quantitative experiments, we use the widely adopted mean Intersection over Union (mean-IoU) metric. The class-wise IoU is calculated as the sum of intersections over union for that class across all samples, and the mean-IoU is obtained by averaging class-wise IoU values across all classes. Reported scores are averaged over 5 independent runs, each corresponding to a different seed from the set $\{0,1,2,3,4\}$. Each run involves 1000 distinct tasks, with each task comprising a randomly selected set of support and query images.

%
\begin{table*}[t]
\vskip 0.2in
\centering
\caption{Few-Shot Binary Segmentation using Bounding-Box Annotations on PASCAL-$5^i$ in terms of mean-IoU.}
\vskip -0.1in
\resizebox{\textwidth}{!}{
\begin{tabular}{cccccc|ccccc|ccccc}
\hline
       & \multicolumn{5}{c}{1-Shot}                & \multicolumn{5}{|c}{5-Shot}  & \multicolumn{5}{|c}{10-Shot} \\
       \cline{2-16} 
Method & Fold-0 & Fold-1 & Fold-2 & Fold-3 & Mean  & Fold-0 & Fold-1 & Fold-2 & Fold-3 & Mean & Fold-0 & Fold-1 & Fold-2 & Fold-3 & Mean \\ \hline
FSBBA-Baseline~\cite{han2023learning} & - & - & - & - & 41.1 & - & - & - & - & 46.4 & - & - & - & - & -\\
FSBBA~\cite{han2023learning} & - & - & - & - & 42.4 & - & - & - & - & 50.0 & - & - & - & - & -\\
RePRI~\cite{boudiaf2021few}  & 18.3  & 26.2  & 36.6  & 29.1  & 27.5 & 20.3  & 28.6  & 40.4  & 31.9  & 30.3 & 20.8 & 29.2 & 41.6 & 32.5 & 31.0\\
BD-CSPN~\cite{liu2020prototype} & \textbf{47.6} & \textbf{35.3} & 50.1 & 42.7 & 43.9 & 46.9 & 35.6 & 54.2 & 47.0 & 45.9 & 45.7 & 35.3 & 55.2 & 47.5 & 45.9\\
SimpleShot~\cite{wang2019simpleshot}  & 40.7   & 34.0  & 48.0  & 43.0 & 41.4  & 43.3  & 34.8  & 53.3  & 47.9 & 44.8 & 43.9 & 34.9 & 54.6 & 48.5 & 45.5\\
ProMi (proposed)  & 47.4  & 35.0  & \textbf{53.9}  & \textbf{45.4}  & \textbf{45.4} & \textbf{53.2}  & \textbf{37.2}  & \textbf{62.7}  & \textbf{52.9}  & \textbf{51.5} & \textbf{56.2} & \textbf{38.5} & \textbf{66.2} & \textbf{55.8} & \textbf{54.2}\\
\hline
\end{tabular}
}
\label{tab:resnet50_results_pascal}
\end{table*}

\begin{table*}[t]
\centering
\caption{Few-Shot Binary Segmentation using Bounding-Box Annotations on COCO-$20^i$ in terms of mean-IoU.}
\vskip -0.1in
\resizebox{\textwidth}{!}{
\begin{tabular}{cccccc|ccccc|ccccc}
\hline
       & \multicolumn{5}{c}{1-Shot}                & \multicolumn{5}{|c}{5-Shot} & \multicolumn{5}{|c}{10-Shot} \\
       \cline{2-16} 
Method & Fold-0 & Fold-1 & Fold-2 & Fold-3 & Mean  & Fold-0 & Fold-1 & Fold-2 & Fold-3 & Mean & Fold-0 & Fold-1 & Fold-2 & Fold-3 & Mean \\ \hline
RePRI~\cite{boudiaf2021few}  & 5.1  & 22.0  & 25.0  & 17.7  & 17.5 & 5.7  & 23.4  & 26.9  & 19.4  & 18.8 & 6.2 & 23.7 & 27.3 & 20.0 & 19.3\\ 
BD-CSPN~\cite{liu2020prototype} & 10.7 & 28.0 & 29.8 & 25.1 & 23.4 & 12.3 & 30.3 & 31.0 & 28.5 & 25.5 & 12.2 & 30.5 & 31.2 & 28.9 & 25.7\\ 
SimpleShot~\cite{wang2019simpleshot}  & 10.2 & 26.6  & 29.7 & 24.7 & 22.8 & 12.3  & 30.2  & 31.1 & 28.4 & 25.5 & 12.3 & 30.5 & 31.4 & 28.9 & 25.7\\ 
ProMi (proposed)  & \textbf{11.4} & \textbf{30.2} & \textbf{33.5} & \textbf{29.5} & \textbf{26.1} & \textbf{13.5} & \textbf{36.6} & \textbf{39.5} & \textbf{35.2} & \textbf{31.2} & \textbf{15.4} & \textbf{39.2} & \textbf{42.3} & \textbf{37.6} & \textbf{33.6}\\
\hline
\end{tabular}
}
\vskip -0.1in
\label{tab:resnet50_results_coco}
\end{table*}

\subsection{Comparative Experiments with Standard CNN-Based Feature Extractors}
\label{subsec_standard_exps}

We conduct few-shot binary segmentation experiments using bounding-box annotations on the PASCAL-$5^{i}$ and COCO-$20^{i}$ datasets, which are respectively built from PASCAL VOC 2012 \cite{everingham2010pascal} and MS-COCO \cite{lin2014microsoft}. To ensure fair comparisons with prior work, each dataset is divided into 4 folds: PASCAL-$5^{i}$ with 20 categories and COCO-$20^{i}$ with 80 categories. For each fold, 15 classes from PASCAL-$5^{i}$ and 60 classes from COCO-$20^{i}$ are used as base classes to pre-train the CNN backbone, while the remaining 5 classes from PASCAL-$5^{i}$ and 20 classes from COCO-$20^{i}$ are used for testing. For each fold, we use a PSPNet~\cite{zhao2017pyramid} with a ResNet-50~\cite{he2016deep} backbone, and pre-train it on the base classes. During testing, the support and query images are resized to a fixed 417 $\times$ 417 resolution and passed through the pre-trained CNN to generate spatial feature maps of size 53$\times$53$\times$512. Despite overlapping receptive fields in the ResNet-50 backbone, we assume that each 1D feature vector corresponds to a distinct pixel patch.

Since Han et al.~\cite{han2023learning} is the only study that evaluated their method using bounding box annotations for few-shot binary segmentation, we primarily compare our approach with the two methods they reported: FSBBA-baseline and FSBBA. Additionally, we evaluated RePRI~\cite{boudiaf2021few}, a few-shot segmentation strategy originally designed for few-shot 1-way (i.e. binary) segmentation with pixel-level annotations in the support set. Here, we directly apply it using bounding box annotations similar to the setting in FSBBA-baseline. To further broaden the comparison, we evaluate robust prototype-based methods SimpleShot~\cite{wang2019simpleshot} and BD-CSPN~\cite{liu2020prototype}, both of which are few-shot image classification strategies that we have adapted to the few-shot segmentation task using bounding box annotations. 

Tables \ref{tab:resnet50_results_pascal} and \ref{tab:resnet50_results_coco} compare ProMi with the other methods on 1-way (one foreground class and one background class), 1-Shot, 5-Shot, and 10-Shot settings, using mean-IoU as the evaluation metric. As the exact class splits are not reported in~\cite{han2023learning}, we only compare against the mean over the 4 folds for the methods FSBBA-baseline and FSBBA reported in~\cite{han2023learning} on PASCAL-$5^{i}$. 
Overall, the results show that despite its simplicity and computational efficiency, ProMi consistently achieves the highest scores compared to other methods.

\subsection{Experiments with the Foundation Model Dinov2}
\label{subsec_foundation_model_exps}
%
In this section, we explore the potential of combining our approach with a more powerful feature extractor: the foundation model Dinov2 \cite{oquab2024dinov}. Due to its generalization capabilities, we do not pre-train it on base classes, allowing direct evaluation on the full PASCAL VOC 2012 \cite{everingham2010pascal} and MS-COCO \cite{lin2014microsoft} datasets without dataset splitting. We use the backbone ViT-B/14 for Dinov2. The support and query images are resized to a fixed 672 $\times$ 672 resolution and passed through Dinov2 to generate spatial feature maps of size 48$\times$48$\times$768. We compare ProMi against adapted versions of SimpleShot~\cite{wang2019simpleshot} and BD-CSPN~\cite{liu2020prototype}, which achieved the most competitive scores after our method in the experiments reported in Sec. \ref{subsec_standard_exps}. 

Table~\ref{tab_fs_using_bbox_annot} shows that ProMi consistently achieves the highest mean-IoU scores across 1-shot, 5-shot, and 10-shot settings on both datasets. On PASCAL VOC 2012, ProMi improves by up to 4.5\% over BD-CSPN in the 1-shot setting and also demonstrates the best performance with increased support data. On MS-COCO, ProMi outperforms BD-CSPN and SimpleShot in the 1-shot setting, with even larger performance gaps in the 5-shot and 10-shot scenarios. These results confirm ProMi's effectiveness in managing complex backgrounds and noisy annotations, and its superior adaptability when paired with a strong foundation model.

%
\begin{table}[t]
\addtolength{\tabcolsep}{-3pt}
\caption{Few-Shot Binary Segmentation with Bounding-Box Labels with Dinov2 (ViT-B/14) pre-training, in terms of mean-IoU.}
\vskip -0.1in
\label{tab_fs_using_bbox_annot}
\centering
    \begin{small}
            \scalebox{0.9}{ %
            \resizebox{\columnwidth}{!}{%
            \begin{tabular}{lcccccc}
                \toprule
                Method & \multicolumn{3}{c}{PASCAL VOC 2012} & \multicolumn{3}{c}{MS-COCO} \\
                \cmidrule(r){2-4} \cmidrule(r){5-7}
                 & 1-Shot & 5-Shot & 10-Shot & 1-Shot & 5-Shot & 10-Shot \\
                 \midrule
                 BD-CSPN~\cite{liu2020prototype} & 39.6 & 44.3 & 45.9 & 18.1 & 22.4 & 23.4 \\
                 SimpleShot~\cite{wang2019simpleshot} & 36.7 & 43.7 & 45.8 & 16.5 & 22.3 & 23.9 \\
                 ProMi (proposed) & \textbf{44.1} & \textbf{50.4} & \textbf{53.2} & \textbf{20.7} & \textbf{28.4} & \textbf{31.2} \\
                \bottomrule
            \end{tabular}
            }
            }
    \end{small}
    \vskip -0.2in
\end{table}

\subsection{Qualitative Experiments for Real-World Mobile Robot Applications}
\label{subsec_qualitative_exps}

\begin{figure*}%
    \vskip 0.15in
    \centering

    \begin{minipage}{0.23\textwidth}%
        \includegraphics[width=\columnwidth]{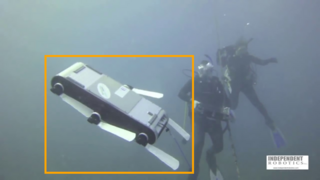}%
        \centering
    \end{minipage}%
    \hspace{0.005\textwidth}
    \begin{minipage}{0.23\textwidth}%
        \includegraphics[width=\columnwidth]{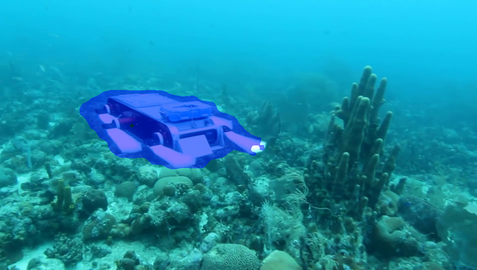}%
        \centering
    \end{minipage}%
    \hspace{0.005\textwidth}
    \begin{minipage}{0.23\textwidth}%
        \includegraphics[width=\columnwidth]{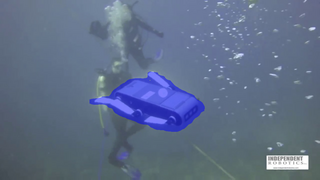}%
        \centering
    \end{minipage}%
    \hspace{0.005\textwidth}
    \begin{minipage}{0.23\textwidth}%
        \includegraphics[width=\columnwidth]{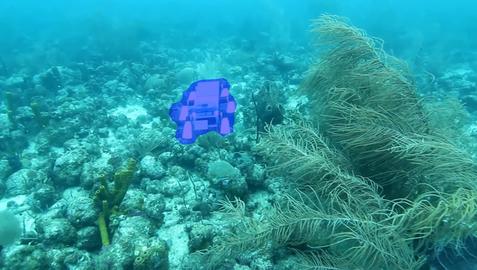}%
        \centering
    \end{minipage}%

    \vspace{0.005\textwidth}
    
    \begin{minipage}{0.23\textwidth}%
        \includegraphics[width=\columnwidth]{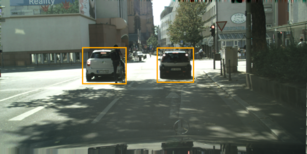}%
        \centering
        \\ (a) One-Shot annotation
    \end{minipage}%
    \hspace{0.005\textwidth}
    \begin{minipage}{0.23\textwidth}%
        \includegraphics[width=\columnwidth]{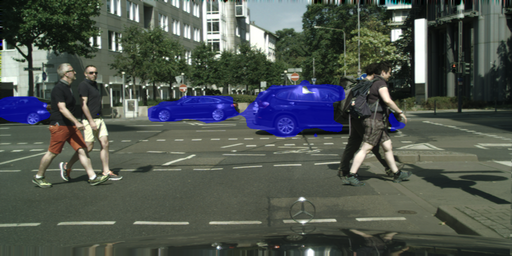}%
        \centering
        \\  \hspace{1cm}%
    \end{minipage}%
    \hspace{0.005\textwidth}
    \begin{minipage}{0.23\textwidth}%
        \includegraphics[width=\columnwidth]{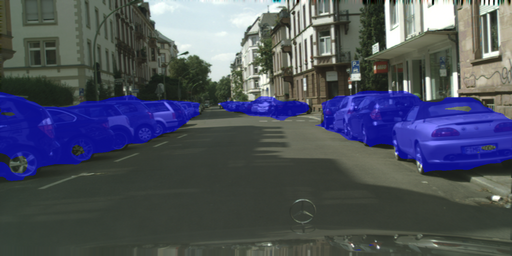}%
        \centering
        \\ (b) Predictions
    \end{minipage}%
    \hspace{0.005\textwidth}
    \begin{minipage}{0.23\textwidth}%
        \includegraphics[width=\columnwidth]{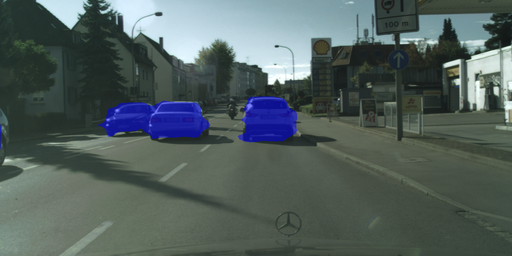}%
        \centering
        \\ \hspace{1cm} %
    \end{minipage}%
    
    \caption{\textbf{Qualitative results on mobile robot datasets.} Our approach was applied in the 1-shot setting to segment the "underwater robot" class on SUIM \cite{islam2020semantic} (first row) and the "car" class on Cityscapes \cite{Cordts2016Cityscapes} (second row). The first column shows 1-shot annotations, followed by three corresponding predictions.}%
    \vspace{-0.015\textwidth}
    \label{fig_qualitative_results_1}%
    \vskip -0.05in
\end{figure*}

\begin{figure}[t] 
    \centering
    \includegraphics[width=0.99\columnwidth]{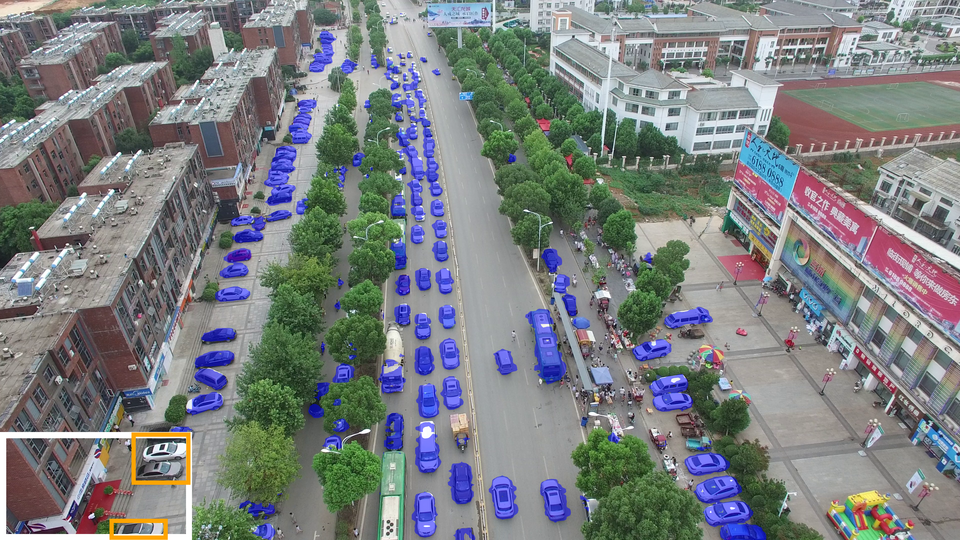} 
    \vskip -0.1in
    \caption{\textbf{Qualitative result on the high-resolution UAVid dataset.} Our method was trained on a small annotated window, outlined in white in the bottom left of the image, and then used to infer segmentation predictions on the remaining areas.} 
    \vskip -0.05in
    \label{fig_uavid_result}
\end{figure}

We perform qualitative experiments on SUIM \cite{islam2020semantic}, Cityscapes \cite{Cordts2016Cityscapes}, and UAVid \cite{lyu2020uavid} 
to demonstrate the versatility of our method in various real-world mobile robotics applications, including underwater \cite{islam2020semantic}, ground \cite{Cordts2016Cityscapes}, and aerial \cite{lyu2020uavid} environments. We provide 1-shot segmentation results using bounding-box annotations, which we manually created for this purpose. We used the Dinov2 encoder with ViT-B/14 backbone as in sec. \ref{subsec_foundation_model_exps}.

Figure \ref{fig_qualitative_results_1} shows ProMi's performance in segmenting "underwater robot" and "car" classes on SUIM and Cityscapes datasets, respectively. The figure includes the initial 1-shot annotations and three segmentation predictions, illustrating ProMi's strong generalization across different backgrounds.

Figure \ref{fig_uavid_result} highlights ProMi's 1-shot segmentation on high-resolution UAVid aerial image (3840 × 2160 pixels). The image is divided into 25 smaller windows to fit the model's input size, with only the bottom-left window annotated. The model then infers pixel labels on the remaining windows, demonstrating its effectiveness in aerial environments.

\subsection{Ablation study}
\label{subsec_ablation_study}

\begin{table}[t]
\centering
\caption{Ablation study of ProMi components on PASCAL VOC 2012 using Dinov2 (ViT-B/14 backbone), in terms of mean-IoU.}
\vskip -0.1in
\scalebox{0.8}{
\resizebox{\columnwidth}{!}{
\begin{tabular}{ccccc}
\hline
BG Mixt. & FG Ref. & 1-Shot & 5-Shot & 10-Shot \\ \hline
$\times$ & $\times$ & 33.4 & 44.2 & 46.5 \\
\checkmark & $\times$ & 40.7 & 46.0 & 48.4 \\
\checkmark & \checkmark & \textbf{44.1} & \textbf{50.4} & \textbf{53.2} \\
\hline
\end{tabular}
}
}
\label{tab_ablation_of_promi}
\vskip -0.2in
\end{table}
\begin{figure}[t] 
    \centering
    \includegraphics[width=0.8\columnwidth]{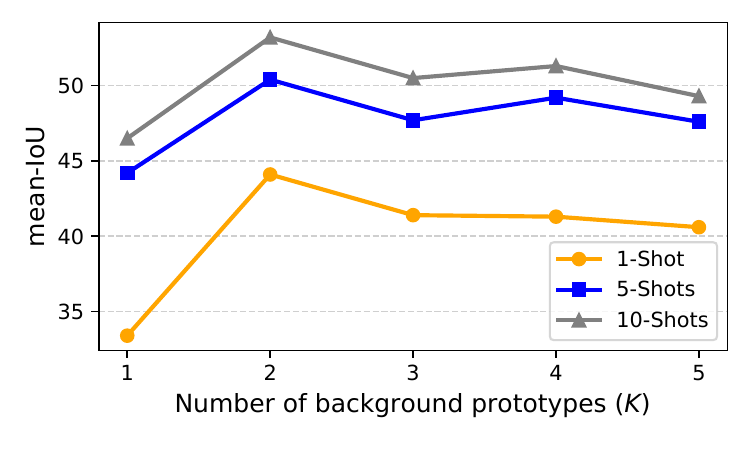} 
    \vskip -0.2in
    \caption{\textbf{ProMi mean-IoU scores as a function of the number of background prototypes}, on PASCAL VOC 2012 using the Dinov2 model (ViT-B/14 backbone).}
    \label{fig_bg_proto_nbr}
    \vskip -0.2in
\end{figure}

\textbf{The number of background prototypes.} Figure \ref{fig_bg_proto_nbr} shows the mean-IoU scores of ProMi on PASCAL VOC 2012 as a function of the number of background prototypes $K$. The performance peaks at $K=2$, before slightly declining and stabilizing, suggesting that using two background prototypes best captures the background distribution variability without adding unnecessary complexity. Therefore, we used two background prototypes in all other experiments. This setting requires far fewer iterations than the tens to hundreds typically needed to train a linear classifier.

\textbf{Ablation study of ProMi components.} Table~\ref{tab_ablation_of_promi} presents the ablation study of ProMi in 1-shot, 5-shot, and 10-shot settings on PASCAL VOC 2012, evaluated in terms of mean-IoU. The pre-trained encoder used is Dinov2 with ViT-B/14 backbone. "BG Mixt." denotes the activation of ProMi's background prototype mixture process with \(K=2\), enabling the model to better capture diverse background variations, as detailed in steps "2)" and "3)" in Sec. \ref{subsec_ProMi}. "FG Ref." indicates the activation of the foreground prototype refinement, as detailed in step "4)" of Sec. \ref{subsec_ProMi}.

\noindent Without any prototype refinement ("BG Mixt." and "FG Ref." both inactive), ProMi achieves a baseline performance. Activating the background prototype mixture ("BG Mixt.") significantly improves performance across all settings. Further activation of foreground prototype refinement ("FG Ref.") leads to the highest performance gains, demonstrating the critical role of both components in enhancing segmentation predictions.

\section{CONCLUSION}
In this paper, we introduce ProMi, a novel training-free method for few-shot binary segmentation using bounding-box annotations. ProMi efficiently adapts to diverse image sets by iteratively adding background prototypes, achieving state-of-the-art performance on multiple benchmark datasets. ProMi also shows promising results on real-world robotic data across aerial, ground, and underwater environments. We hope this work advances visual sensing and facilitates the deployment of robots in diverse real-world applications.

\textbf{Limitations and future work.} While our method is training-free and highly efficient, it currently focuses on binary segmentation and relies on manual bounding-box annotations in the support set. Future work could explore multi-class segmentation and automated annotation strategies, particularly in scenarios where robots encounter novel objects introduced dynamically, enabling real-time interaction and adaptive learning.

{\small
\bibliographystyle{ieee_fullname}
\bibliography{egbib}
}

\end{document}